\def\year{2020}\relax
\documentclass[letterpaper]{article} 

\newif\ifaaai
\aaaifalse
\usepackage{aaai20}  
\usepackage{times}  
\usepackage{helvet} 
\usepackage{courier}  
\usepackage[hyphens]{url}  
\usepackage{graphicx} 
\usepackage{graphicx}  
\usepackage{url}
\usepackage{xspace}
\usepackage{xcolor}
\usepackage{colortbl}
\usepackage{tcolorbox}
\usepackage{tikz}
\usepackage{multirow}
\usepackage{enumitem}
\usepackage{amsmath,amssymb}
\usepackage{footmisc}
\usepackage{soul}

\newcommand{\ai}{}
\newcommand{\ua}{$^+$}

\setul{0.5ex}{0.4ex}

\newcommand{\dataset}{\textsf{QASC}\xspace}
\newcommand{\datasetfullform}{\textbf{Q}uestion \textbf{A}nswering via \textbf{S}entence \textbf{C}omposition\xspace}
\newcommand{\datasetpronounce}{kask}
\newcommand{\datasetSize}{9,980\xspace}

\newcommand{\uldots}[1]{%
    \tikz[baseline=(todotted.base)]{
        \node[inner sep=1pt,outer sep=0pt] (todotted) {#1};
        \draw[loosely dotted] (todotted.south west) -- (todotted.south east);
    }%
}%
\newcommand{\uldashes}[1]{%
    \tikz[baseline=(todotted.base)]{
        \node[inner sep=1pt,outer sep=0pt] (todotted) {#1};
        \draw[loosely dashed] (todotted.south west) -- (todotted.south east);
    }%
}%
\newcommand{\ulsolid}[1]{%
    \tikz[baseline=(todotted.base)]{
        \node[inner sep=1pt,outer sep=0pt] (todotted) {#1};
        \draw[solid] (todotted.south west) -- (todotted.south east);
    }%
}%

\ifaaai
\definecolor{colgreen}{cmyk}{0,0,0,0.1}
\definecolor{colblue}{cmyk}{0,0,0,0.3}
\definecolor{colred}{cmyk}{0,0,0,0.8}
\newcommand{\colredname}{black}
\else
\definecolor{colgreen}{HTML}{EEFFDD}
\definecolor{colblue}{HTML}{DDEEFF}
\definecolor{colred}{HTML}{E06666}
\newcommand{\colredname}{red}
\fi
\newcommand{\cbox}[2]{{\setlength{\fboxsep}{0.5pt}\colorbox{#1}{#2}}}
\newcommand{\hgreen}[1]{\uldots{\cbox{colgreen}{#1}}}
\newcommand{\hblue}[1]{{\uldashes{\cbox{colblue}{#1}}}}
\newcommand{\hred}[1]{\ulsolid{\textbf{\color{white}\cbox{colred}{#1}}}}

\newcommand{\cboxr}[2]{
  \begin{tcolorbox}[colback=#1,boxrule=0.5pt,hbox,top=0.5pt,left=0.5pt, right=0.5pt,bottom=0.5pt,boxsep=0.5pt,nobeforeafter]{#2}\end{tcolorbox}}

\newcommand{\FS}{\ensuremath{F_S}\xspace} 
\newcommand{\facttext}[1]{\textsl{``{#1}''}}

\newcommand{\FL}{\ensuremath{F_L}\xspace} 
\newcommand{\fS}{\ensuremath{f_S}\xspace} 
\newcommand{\fL}{\ensuremath{f_L}\xspace} 
\newcommand{\fc}{\ensuremath{f_C}\xspace} 
\newcommand{\FSL}{\ensuremath{F_{\textit{QASC}}}\xspace} 
\newcommand{\mk}{\ensuremath{m_k}\xspace}

\newcommand{\disr}{\ensuremath{D}}
\newcommand{\disi}{\ensuremath{d_i}\xspace}
\newcommand{\rQ}{\ensuremath{r_Q}\xspace} 
\newcommand{\rS}{\ensuremath{r_S}\xspace} 
\newcommand{\rL}{\ensuremath{r_L}\xspace} 

\newcommand{\embboxc}[2]{\cboxr{#1}{\small {#2}}}

\ifaaai

\newcommand{\glove}{\embboxc{gray!15!white}{Glove}}
\newcommand{\elmo}{\embboxc{gray!35!white}{Elmo}}
\newcommand{\embbox}[1]{\cboxr{gray!50!white}{\small {#1}}}

\else
\newcommand{\embbox}[1]{\cboxr{yellow!50!white}{\small {#1}}}

\newcommand{\glove}{\embboxc{gray!50!white}{Glove}}
\newcommand{\elmo}{\embboxc{red!50!white}{Elmo}}

\fi
\newcommand{\bertmcq}{BERT-MCQ\xspace}
\newcommand{\bertlc}{\embbox{BERT-LC}}

\newcommand{\bertwm}{\embbox{BERT-LC[WM]}}

\newcommand{\race}{\ensuremath{\textsc{Race}}\xspace}
\newcommand{\sci}{\ensuremath{\textsc{Sci}}\xspace}

\newcommand{\twostep}{two-step\xspace}
\newcommand{\Twostep}{Two-step\xspace}

\newcommand{\onestep}{single-step\xspace}
\newcommand{\Onestep}{Single-step\xspace}

\newcommand{\ques}{\ensuremath{q}\xspace}

\newcommand{\ans}{\ensuremath{a}\xspace}

\newcommand{\choicei}{\ensuremath{c_i}\xspace}
\newcommand{\parai}{\ensuremath{p_i}\xspace}

\newcommand\T{\rule{0pt}{2.6ex}}       
\newcommand\B{\rule[-1.2ex]{0pt}{0pt}} 
\newcommand\R{\rule{0pt}{2.0ex}}       

\newcommand{\namecite}[1]{\citeauthor{#1}~\shortcite{#1}}

\title{QASC: A Dataset for Question Answering via Sentence Composition}

 \author{
 Tushar Khot\ai \and Peter Clark\ai \and Michal Guerquin\ai \and Peter Jansen\ua \and Ashish Sabharwal\ai\\
 \\
  \ai Allen Institute for AI,
  Seattle, WA, U.S.A.\\
   \ua University of Arizona, Tucson, AZ, U.S.A. \\
   \\
   {\small \{tushark,peterc,michalg,ashishs\}@allenai.org, pajansen@email.arizona.edu}
 }
 
\date{}

\begin{document}

\maketitle


\begin{abstract}
Composing knowledge from multiple pieces of texts is a key challenge in multi-hop question answering. We present a multi-hop reasoning dataset, \datasetfullform (\dataset), that requires retrieving facts from a large corpus and composing them to answer a multiple-choice question. \dataset is the first dataset to offer two desirable properties: (a) the facts to be composed are annotated in a large corpus, and (b) the decomposition into these facts is not evident from the question itself. The latter makes retrieval challenging as the system must introduce new concepts or relations in order to discover potential decompositions. Further, the reasoning model must then learn to identify valid compositions of these retrieved facts using common-sense reasoning. To help address these challenges, we provide annotation for supporting facts as well as their composition. Guided by these annotations, we present a two-step approach to mitigate the retrieval challenges. We use other multiple-choice datasets as additional training data to strengthen the reasoning model. Our proposed approach improves over current state-of-the-art language models by 11\% (absolute). The reasoning and retrieval problems, however, remain unsolved as this model still lags by 20\% behind human performance.

\end{abstract}

\section{Introduction}
\label{sec:intro}

\begin{figure}[t]
\fbox{
\begin{minipage}{0.45\textwidth}
\small
\textbf{Question:} 
\hgreen{Differential heating of air} can be harnessed for what?\\
\noindent\begin{tabular}{@{}ll}
(A) \hblue{electricity production} & (D) reduce acidity of food \\
(B) erosion prevention             & $\ldots$ \\
(C) transfer of electrons          & $\ldots$ \\
\end{tabular}\\
\noindent\rule{\textwidth}{0.5pt}\\
\textbf{Annotated facts:}

\T \fS: \hgreen{Differential heating of air} produces \hred{wind}.

\T \fL: \hred{Wind} is used for \hblue{producing electricity}.

\T Composed fact \fc: \hgreen{Differential heating of air} can be harnessed for \hblue{electricity production}.

\end{minipage}
}
\caption{
  \label{figure:dataset-example}
  A sample 8-way multiple choice \dataset question.
  Training data includes the associated facts \fS and \fL shown above, as well as their composition \fc.
  The term \hred{wind} connects \fS and \fL, but appears neither in \fc nor in the question.
  Further, decomposing the question relation \facttext{harnessed for} into \fS and \fL requires introducing the new relation \facttext{produces} in \fS.
  The question can be answered by using broad knowledge to compose these facts together and infer \fc.
}
\end{figure}

Several multi-hop question-answering (QA) datasets have been proposed to promote research on multi-sentence machine comprehension. On one hand, many of these datasets~\cite{dataset:openbookqa,ARCClark2018,wikihop,Talmor2018TheWA} do not come annotated with sentences or documents that can be combined to produce an answer. Models must thus learn to reason without direct supervision. On the other hand, datasets that come with such annotations involve either single-document questions~\cite{MultiRCKhashabi2018} leading to a strong focus on coreference resolution and entity tracking, or multi-document questions~\cite{dataset:hotpotqa} whose decomposition into simpler single-hop queries is often evident from the question itself.

We propose a novel dataset, \datasetfullform (\dataset; pronounced \datasetpronounce) of \datasetSize multi-hop multiple-choice questions (MCQs) where simple syntactic cues are insufficient to determine how to decompose the question into simpler queries. Fig.~\ref{figure:dataset-example} gives an example, where the question is answered by decomposing its main relation \facttext{harnessed for} (in \fc) into a similar relation \facttext{used for} (in \fL) and a newly introduced relation \facttext{produces} (in \fS), and then composing these back to infer \fc.

\ifaaai
\else
    \newcommand{\yes}{{\cellcolor{green!15}Y}}
    \newcommand{\no}{{\cellcolor{yellow!25}N}}
    \newcommand{\sometimes}{{\cellcolor{gray!15}?}}
    \newcommand{\unclear}{{\cellcolor{gray!15}--}}
\fi

\begin{table*}
\small
\centering
\setlength{\tabcolsep}{5pt}
\setlength{\doublerulesep}{\arrayrulewidth}
\begin{tabular}{l||c|c|c|c|c|c||c}
\B Property & CompWebQ & DROP & HotPotQA & MultiRC & OpenBookQA & WikiHop & \textbf{\dataset} \\
\hline \hline
\T Supporting facts are available     &  \no & \yes &  \yes & \yes & \no & \no  & \yes \\
Supporting facts are annotated     &  \no &  \no & \yes & \yes  & \no  & \no  & \yes \\
Decomposition is not evident & \no & \unclear & \no & \yes & \yes & \yes & \yes \\
Multi-document inference & \yes & \no & \no & \no & \yes & \no & \yes \\
Requires knowledge retrieval & \yes & \no & \yes & \no & \yes & \no & \yes \\
\end{tabular}
\caption{
  \label{table:dataset-comparison}
  \dataset has several desirable properties not simultaneously present in any single existing multihop QA dataset. Here ``available'' indicates that the dataset comes with a corpus that is guaranteed to contain supporting facts, while ``annotated'' indicates that these supporting facts are additionally annotated. 
}
\end{table*}


While the question in Figure~\ref{figure:dataset-example} can be answered by composing the two facts \fS and \fL, that this is the case is unclear based solely on the question. This property of relation decomposition not being evident from reading the question pushes reasoning models towards focusing on learning to compose new pieces of knowledge, a key challenge in language understanding. Further, $\fL$ has no overlap with the question, making it difficult to retrieve it in the first place.

Let's contrast this with an alternative question formulation: 
``What can something \emph{produced by} differential heating of air be \emph{used for}?''
Although awkwardly phrased, this variation is easy to syntactically decompose into two simpler queries, as well as to identify what knowledge to retrieve. In fact, multi-hop questions in many existing datasets~\cite{dataset:hotpotqa,Talmor2018TheWA} often follow this syntactically decomposable pattern, with questions such as: ``Which government position was held by the lead actress of X?''

All questions in \dataset are human-authored, obtained via a multi-step crowdsourcing process (Section~\ref{sec:collection}). To better enable development of both the reasoning and retrieval models, we also provide the pair of facts that were composed to create the question.\footnote{Questions, annotated facts, and corpora are available at https://github.com/allenai/qasc. } We use these annotations to develop a novel \emph{two-step retrieval} technique that uses question-relevant facts to guide a second retrieval step. To make the dataset difficult for fine-tuned language models using our proposed retrieval (Section~\ref{sec:models}), we further augment the answer choices in our dataset via a \emph{multi-adversary distractor choice selection} method (Section~\ref{sec:adversarial}) that does not rely on computationally expensive multiple iterations of adversarial filtering~\cite{dataset:swag}.

Even 2-hop reasoning for questions with implicit decomposition requires new approaches for retrieval and reasoning not captured by current datasets. Similar to other recent multi-hop reasoning tasks~\cite{dataset:hotpotqa,Talmor2018TheWA}, we also focus on 2-hop reasoning, solving which will go a long way towards more general N-hop solutions.

In summary, we make the following contributions: (1) a dataset \dataset of \datasetSize 8-way multiple-choice questions from elementary and middle school level science, with a focus on fact composition; (2) a pair of facts \fS,\fL from associated corpora annotated for each question, along with a composed fact \fc entailed by \fS and \fL, which can be viewed as a form of multi-sentence entailment dataset;
(3) a novel \twostep information retrieval approach designed for multi-hop QA that improves the recall of gold facts (by 43 pts) and QA accuracy (by 14 pts); and
(4) an efficient multi-model adversarial answer choice selection approach.

\dataset is challenging for current large pre-trained language models~\cite{elmo,bert}, which exhibit a gap of 20\% (absolute) to a human baseline of 93\%, even when massively fine-tuned on 100K external QA examples in addition to \dataset and provided with relevant knowledge using our proposed \twostep retrieval.

\section{Comparison With Existing Datasets}

Table~\ref{table:dataset-comparison} summarizes how \dataset compares with several existing datasets along five key dimensions (discussed below), which we believe are necessary for effectively developing retrieval and reasoning models for knowledge composition.

Existing datasets for the science domain require different reasoning techniques for each question~\cite{clark2016combining,ARCClark2018}. The dataset most similar to our work is OpenBookQA~\cite{dataset:openbookqa}, which comes with multiple-choice questions and a book of core science facts used as the seed for question generation. Each question requires combining the seed core fact with additional knowledge. However, it is unclear how many additional facts are needed, or whether these facts can even be retrieved from any existing knowledge sources. \dataset, on the other hand, explicitly identifies two facts deemed (by crowd workers) to be sufficient to answer a question. These facts exist in an associated corpus and are provided for model development. 

MultiRC~\cite{MultiRCKhashabi2018} uses passages
to create multi-hop questions. However, MultiRC and other single-passage datasets~\cite{Propara:Mishra2018TrackingSC,babi-Weston-15} have a stronger emphasis on passage discourse and entity tracking, rather than relation composition.

Multi-hop datasets from the Web domain use complex questions that bridge multiple sentences. We discuss 4 such datasets. (a) WikiHop~\cite{wikihop} contains questions in the tuple form $(e, r, ?)$ based on edges in a knowledge graph.
However, WikiHop lacks questions with natural text or annotations on the passages that could be used to answer these questions. (b) ComplexWebQuestions~\cite{Talmor2018TheWA} was derived by converting multi-hop paths in a knowledge-base into a text query. By construction, the questions can be decomposed into simpler queries corresponding to knowledge graph edges in the path.  (c) HotPotQA~\cite{dataset:hotpotqa} contains a mix of multi-hop questions authored by crowd workers using a pair of Wikipedia pages. While these questions were authored in a similar way, due to their domain and task setup, they also end up being more amenable to decomposition. (d) A recent dataset, DROP~\cite{Dua2019DROP}, requires discrete reasoning over text (such as counting or addition).
Its focus is on performing discrete (e.g., mathematical) operations on extracted pieces of information, 
unlike our proposed sentence composition task.

Many systems answer science questions by composing multiple facts from semi-structured and unstructured knowledge sources~\cite{tableilp2016,Khot2017AnsweringCQ,Jansen2017FramingQA,semanticilp2018aaai}. However, these often require careful manual tuning due to the large variety of reasoning techniques needed for these questions~\cite{boratko-etal-2018-systematic} and the large number of facts that often must be composed together~\cite{Jansen2018MultihopIF,Jansen2016Explanation}. By limiting \dataset to require exactly 2 hops (thereby avoiding semantic drift issues with longer paths~\cite{Fried2015HigherorderLS,Khashabi2019Capabilities}) and explicitly annotating these hops, we hope to constrain the problem enough so as to enable the development of supervised models for identifying and composing relevant knowledge.

\subsection{Implicit Relation Decomposition}

As mentioned earlier, a key challenge in \dataset is that syntactic cues in the question are insufficient to determine how one should decompose the question relation, \rQ, into two sub-relations, \rS and \rL, corresponding to the associated facts \fS and \fL. At an abstract level, 2-hop questions in \dataset generally exhibit the following form:
\begin{gather*}
    Q \, \triangleq \, \rQ(x_q, z_a^?)\\
    \rS^?(x_q, y^?) \, \wedge \, \rL^?(y^?, z_a^?) \Rightarrow \rQ(x_q, z_a^?)
\end{gather*}
where terms with a `?' superscript represent unknowns: the decomposed relations \rS and \rL as well as the bridge concept $y$. (The answer to the question, $z_a^?$, is an obvious unknown.)
To assess whether relation \rQ holds between some concept $x_q$ in the question and some concept $z_a$ in an answer candidate, one must come up with the missing or implicit relations and bridge concept. In our previous example, $\rQ=$\facttext{harnessed for}, $x_q=$\facttext{Differential heating of air}, $y=$\facttext{wind}, $\rS=$\facttext{produces}, and $\rL=$\facttext{used for}.

In contrast, syntactically decomposable questions in many existing datasets often spell out both \rS and \rL: $ Q \, \triangleq \, \rS(x_q, y^?) \, \wedge \, \rL(y^?, z_a^?)$.
%
%
The example from the introduction, ``Which government position was held by the lead actress of X?'', could be stated in this notation as:
$
    \textit{lead-actress}(\text{X}, y^?) \, \wedge \, \textit{held-govt-posn}(y^?, z_a^?).
$

This difference in how the question is presented in \dataset makes it challenging to both retrieve relevant facts and reason with them via knowledge composition. This difficulty is further compounded by the property that a single relation \rQ can often be decomposed in \emph{multiple ways} into \rS and \rL. We defer a discussion of this aspect to later, when describing \dataset examples in Table~\ref{tab:composition-examples}.

\section{Multihop Question Collection}
\label{sec:collection}

\begin{figure}[t]
        \centering
        \includegraphics[width=0.28\textwidth]{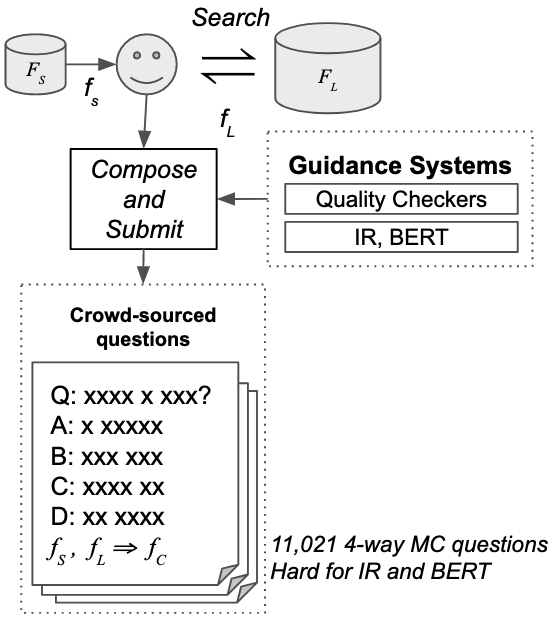}
        \caption{
            \label{fig:dataset-crowdsourcing}  Crowd-sourcing questions using the seed corpus \FS and the full corpus \FL.
        }
        
\end{figure}

Figure~\ref{fig:dataset-crowdsourcing} gives an overall view of the crowdsourcing process. The process is designed such that each question in \dataset is produced by composing two facts from an existing text corpus. Rather than creating compositional questions from scratch or using a specific pair of facts, we provide workers with only one seed fact \fS as the starting point. They are then given the creative freedom to find other relevant facts from a large corpus, \FL that could be composed with this seed fact. This allows workers to find other facts compose naturally with \fS and thereby prevent complex questions that describe the composition explicitly.  

Once crowd-workers identify a relevant fact \fL$\in$\FL that can be composed with \fS, they create a new composed fact \fc and use it to create a multiple-choice question. To ensure that the composed facts and questions are consistent with our instructions, we introduce automated checks to catch any inadvertent mistakes. E.g., we require that at least one intermediate entity (marked in \hred{\colredname} in subsequent sections) must be dropped to create \fc. We also ensure that the intermediate entity wasn't re-introduced in the question. 

These questions are next evaluated against baseline systems to ensure hardness, i.e., at least one of the incorrect answer choices had to be be preferred over the correct choice by one of two QA systems (IR or BERT; described next), with a bonus incentive if both systems were distracted. 

\subsection{Input Facts}
\label{subsec:facts}

\paragraph{Seed Facts, \FS:} We noticed that the quality of the seed facts can have a strong correlation with the quality of the question. So we created a small set of 928 good quality seed facts \FS from clean knowledge resources. We start with two medium size corpora of grade school level science facts: the WorldTree corpus~\cite{Jansen2018WorldTreeAC} and a collection of facts from the CK-12 Foundation.\footnote{\url{https://www.ck12.org}} Since the WorldTree corpus contains only facts covering elementary science, we used their annotation protocol to
expand it to middle-school science. We then manually selected facts from these three sources that are amenable to creating 2-hop questions.\footnote{While this is a subjective decision, it served our main goal of identifying a reasonable set of seed facts for this task.} The resulting corpus \FS contains a total of 928 facts: 356 facts from WorldTree, 123 from our middle-school extension, and 449 from CK-12.
 
\paragraph{Large Text Corpus, \FL:} To ensure that the workers are able to find any potentially composable fact, we used a large web corpus of 17M cleaned up facts \FL. We processed and filtered a corpus of 73M web documents (281GB) from ~\namecite{clark2016combining} to produce this clean corpus of 17M sentences (1GB). The procedure to process this corpus involved using \textsl{spaCy}\footnote{\url{https://spacy.io/}} 
to segment documents into sentences, a Python implementation of Google's 
\textsl{langdetect}\footnote{\url{https://pypi.org/project/spacy-langdetect/}} to identify
English-language sentences, \textsl{ftfy}\footnote{\url{https://github.com/LuminosoInsight/python-ftfy}}
to correct Unicode encoding problems, and custom heuristics to exclude sentences
with artifacts of web scraping like HTML, CSS and JavaScript markup,
runs of numbers originating from tables, email addresses, URLs, page navigation
fragments, etc.

\subsection{Baseline QA Systems}
\label{subsec:baselines}

Our first baseline is the \textbf{IR} system~\cite{clark2016combining} designed for science QA with its associated corpora of web and science text (henceforth referred as the Aristo corpora). It retrieves sentences for each question and answer choice from the associated corpora, and returns the answer choice with the highest scoring sentence (based on the retrieval score). 


Our second baseline uses the language model \textbf{BERT} of \namecite{bert}. We follow their QA approach for the multiple-choice situation inference task SWAG~\cite{dataset:swag}. Given question $\ques$ and an answer choice $\choicei$, we create \texttt{[CLS] \ques [SEP] \choicei [SEP]} as the input to the model, with \ques being assigned to segment 0 and \choicei to segment 1.\footnote{We assume familiarity with BERT's notation such as \texttt{[CLS]}, \texttt{[SEP]}, uncased models, and masking~\cite{bert}.} The model learns a linear layer to project the representation of the \texttt{[CLS]} token to a score for each choice \choicei. We normalize the scores across all answer choices using softmax and train the model using the cross-entropy loss. When context/passage is available, we append the passage to segment 0, i.e., given a retrieved passage \parai, we provide \texttt{[CLS] \parai \ques [SEP] \choicei [SEP]} as the input. We refer to this model as \bertmcq in subsequent sections.

For the crowdsourcing step, we use the \texttt{bert-large-uncased} model and fine-tuned it sequentially on two datasets: (1) RACE~\cite{Lai2017RACELR} with context; (2) \sci questions (ARC-Challenge+Easy~\cite{ARCClark2018} + OpenBookQA~\cite{dataset:openbookqa} + Regents 12th Grade Exams\footnote{http://www.nysedregents.org/livingenvironment}).

\subsection{Question Validation}
\label{subsec:validation}

We validated these questions by having 5 crowdworkers answer them. Any question answered incorrectly or considered unanswerable by at least 2 workers was dropped, reducing the collection to 7,660 questions. The accuracy of the IR and BERT models used in Step 4 was 32.25\% and 38.73\%, resp., on this reduced subset.\footnote{The scores are not 0\% as crowdworkers were not required to distract both systems for every question.} By design, every question has the desirable property of being annotated with two sentences from \FSL that can be composed to answer it. The low score of the IR model also suggests that these questions can not be answered using a single fact from the corpus.

We next analyze the retrieval and reasoning challenges associated with these questions. Based on these analyses, we will propose a new baseline model for multi-hop questions that substantially outperforms existing models on this task. We use this improved model to adversarially select additional distractor choices to produce the final \dataset dataset.

\section{Challenges}

\begin{table*}[ht]
    \centering
    \small
    \begin{tabular}{p{3.5cm} p{5cm} p{7.5cm}}
    Question & Choices & Annotated Facts \\
    \hline
    What can \textit{trigger immune response}? & \parbox[t]{5cm}{(A) \textbf{Transplanted organs} \\ (B) Desire  \\ (C) Pain \\ (D) Death} & \parbox[t]{7.5cm}{\fS: \hred{Antigens} are found on cancer cells and the cells of \textbf{transplanted organs}.\\ \fL: Anything that can \textit{trigger an immune response} is called an \hred{antigen}.}\\
    What \textit{forms caverns by seeping through rock and dissolving limestone}? & \parbox[t]{4.8cm}{(A)\textbf{ carbon dioxide in groundwater} \\ (B) oxygen in groundwater  \\ (C) pure oxygen \\ (D) magma in groundwater}  & \parbox[t]{7.5cm}{\fS: \textit{a cavern is formed by} \hred{carbonic acid} in \textbf{groundwater} \textit{seeping through rock and dissolving limestone}.\\ \fL: When \textbf{carbon dioxide} is in water, it creates \hred{carbonic acid}.}\\
    \end{tabular}
    \caption{
        \label{tab:question-examples}Examples of questions generated via the crowd-sourcing process along with the facts used to create each question. 
    }
\end{table*}

\newcolumntype{a}{>{\columncolor[gray]{0.8}}c}
\begin{table*}[ht]
    \centering
    \small
    \begin{tabular}{p{4.5cm} a p{4.5cm} a p{3.5cm} a}
    Fact 1 & $\rS$ & Fact 2 & $\rL$ & Composed Fact & $\rQ$ \\
    \hline
    Antigens are found on cancer cells and the cells of transplanted organs. & located & Anything that can trigger an immune response is called an antigen. & causes & transplanted organs can trigger an immune response & causes \\
    \T a cavern is formed by carbonic acid in groundwater seeping through rock and dissolving limestone & causes & Any time water and carbon dioxide mix, carbonic acid is the result. & causes & carbon dioxide in groundwater creates caverns & causes \\
    \end{tabular}
    \caption{
        \label{tab:composition-examples}
        These examples of sentence compositions result in the same composed relation, \texttt{causes}, but via two different composition rules: located + causes $\Rightarrow$ causes and causes + causes $\Rightarrow$ causes
         These rules are not evident from the composed fact, requiring a model reasoning about the composed fact to learn the various possible decompositions of \texttt{causes}.
    }
\end{table*}

Table~\ref{tab:question-examples} shows sample crowd-sourced questions along with the associated facts. Consider the first question: \facttext{What can trigger immune response?}. One way to answer it is to first retrieve the two annotated facts (or similar facts) from the corpus. But the first fact, like many other facts in the corpus, overlaps only with the words in the answer \facttext{transplanted organs} and not with the question, making retrieval challenging. Even if the right facts are retrieved, the QA model would have to know how to compose the \facttext{found on} relation in the first fact with the \facttext{trigger} relation in the second fact. Unlike previous datasets~\cite{dataset:hotpotqa,Talmor2018TheWA}, the relations to be composed are not explicitly mentioned in the question, making reasoning also challenging. We next discuss these two issues in detail.


\subsection{Retrieval Challenges}
We analyze the retrieval challenges associated with finding the two supporting facts associated with each question. Note that, unlike OpenBookQA, we consider the more general setting of retrieving relevant facts from a single large corpus $\FSL=\FS\cup\FL$ instead of assuming the availability of a separate small book of facts (i.e., \FS).

Standard IR approaches for QA retrieve facts using question + answer as their IR query~\cite{clark2016combining,Khot2017AnsweringCQ,semanticilp2018aaai}. While this can be effective for lookup questions, it is likely to miss important facts needed for multi-hop questions. In 96\% of our crowd-sourced questions, at least one of the two annotated facts had an overlap of fewer than 3 tokens (ignoring stop words) with this question + answer query, making it difficult to retrieve such facts.\footnote{\label{footnote:dataset_overlap}See Table~\ref{tab:dataset_overlap} in Appendix~\ref{appendix:overlap_stats} \xspace for more details.} Note that our annotated facts form one possible pair that could be used to answer the question. While retrieving these specific facts isn't necessary, these crowd-authored questions are generally expected to have a similar overlap level to other relevant facts in our corpus.

Neural retrieval methods that use distributional representations can help mitigate the brittleness of word overlap measures, but also vastly open up the space of possibly relevant sentences. We hope that our annotated facts will be useful for training better neural retrieval approaches for multi-hop reasoning in future work. In this work, we focused on a modified non-neural IR approach that exploits the intermediate concepts not mentioned in the question (\hred{\colredname} words in our examples), which is explained in Section~\ref{subsec:two-step}.

\subsection{Reasoning Challenges}
As described earlier, we collected these questions to require compositional reasoning where the relations to be composed are not obvious from the question. To verify this, we analyzed 50 questions from our final dataset and identified the key relations in $\fS, \fL$, and the question, referred to as $\rS, \rL,$ and $\rQ$, respectively (see examples in Table~\ref{tab:composition-examples}). 7 of the 50 questions could be answered using only one fact and 4 of them didn't use either of the two facts. We analyzed the remaining 39 questions to categorize the associated reasoning challenges. In only 2 questions, the two relations needed to answer the question were explicitly mentioned in the question itself. In comparison, the composition questions in HotpotQA had both the relations mentioned in 47 out of 50 dev questions in our analysis. 

Since there are a large number of lexical relations, we focus on 16 semantic relations in our analysis such as \texttt{causes}, \texttt{performs}, etc. These relations were defined based on previous analyses on science datasets~\cite{clark2014automatic,Jansen2016Explanation,tableilp2016}. We found 25 unique relation composition rules (i.e., $\rS(X, Y), \rL(Y, Z) \Rightarrow \rQ(X, Z)$). On average, we found every query relation $\rQ$ had 1.6 unique relation compositions. Table~\ref{tab:composition-examples} illustrates two different relation compositions that lead to the same \texttt{causes} query relation. As a result, models for \dataset have a strong incentive to learn various possible compositions that lead to the same semantic relation, as well as extract them from text.

\section{Question Answering Model}
\label{sec:models}

We now discuss our proposed two-step retrieval method and how it substantially boosts the performance of BERT-based QA models on crowd-sourced questions. This will motivate the need for adversarial choice generation.

\subsection{Retrieval: \Twostep IR}
\label{subsec:two-step}

Consider the first question in Table~\ref{tab:question-examples}. An IR approach that uses the standard $\ques+\ans$ query is unlikely to find the first fact since many irrelevant facts would also have the same overlapping words -- ``transplanted organs''. However, it is likely to retrieve facts similar to the second fact, i.e., \facttext{Antigens trigger immune response}. If we could recognize  \hred{antigen} as an important intermediate entity that would lead to the answer, we can then query for sentences connecting this intermediate entity (\facttext{antigens}) to the answer (\facttext{transplanted organs}) which is then likely to find the first fact (\facttext{antigens are found on transplanted organs}). One potential way to identify such an intermediate concept is to consider the new entities introduced in the first retrieved fact that are absent from the question, i.e., $f_1 \setminus \ques$. 

Based on this intuition, we present a simple but effective \twostep IR baseline for multi-hop QA:
    \textbf{(1)} Retrieve K (=20 for efficiency) facts $F_1$ based on the query Q=\ques + \ans;
    \textbf{(2)} For each $f_1 \in F_1$, retrieve L (=4 to promote diversity) facts $F_2$ each of which contains at least one word from $Q \setminus f_1$ and from $f_1 \setminus Q$;
    \textbf{(3)} Filter $\{f_1, f_2\}$ pairs that do not contain any word from \ques or \ans;
    \textbf{(4)} Select top M unique facts from $\{f_1, f_2\}$ pairs sorted by the sum of their individual IR score.

Each retrieval query is run against an ElasticSearch\footnote{https://www.elastic.co} index built over $\FSL$ with retrieved sentences filtered to reduce noise~\cite{ARCClark2018}. We use the set-difference between the stemmed, non-stopword tokens in $\ques+\ans$ and $f_1$ to identify the intermediate entity. Generally, we are interested in finding facts that connect new concepts introduced in the first fact (i.e., $f_1 \setminus Q$) to concepts not yet covered in question+answer (i.e., $Q \setminus f_1$).

Training a model on our annotations or essential terms~\cite{essentialterms} could help better identify these concepts. Recently, \namecite{Khot2019WhatsMA} proposed a span-prediction model to identify such intermediate entities for OpenBookQA questions. Their approach, however, assumes that one of the gold facts is provided as input to the model. Our approach, while specifically designed for 2-hop questions, can serve as a stepping stone towards developing retrieval methods for N-hop questions.

The single step retrieval approach (using only $f_1$ but still requiring overlap with \ques and \ans) has an overall recall of only 2.9\% (i.e., both $f_S$ and $f_L$ were in the top 10 sentences for 2.9\% of the questions). The \twostep approach, on the other hand, has a recall of 44.4\%---a \textbf{15X improvement} (also limited to top M=10 sentences). Even if we relax the recall metric to finding \fS or \fL, the single step approach under-performs by 28\% compared to the \twostep retrieval (42.0 vs 69.9\%). We will show in the next section that this improved recall also translates to improved QA scores. This shows the value of our \twostep approach as well as the associated annotations: progress on the retrieval sub-task enabled by our fact-level annotations can lead to progress on the QA task.

\subsection{Reasoning: BERT Models}

We primarily use BERT-models fine-tuned on other QA datasets and with retrieved sentences as context, similar to prior state-of-the-art models on MCQ datasets~\cite{Sun2018ImprovingMR,Pan2019ImprovingQA}.\footnote{Experiments section contains numbers for other QA models.} There is a large space of possible configurations to build such a QA model (e.g., fine-tuning datasets, corpora) which we will explore later in our experimental comparisons. For simplicity, the next few sections will focus on one particular model: the \texttt{bert-large-cased} model fine-tuned on the \race + \sci questions (with retrieved context\footnote{We use the same \onestep retrieval over the large Aristo corpus as used by other BERT-based systems on ARC and OpenBookQA leaderboards.}) and then fine-tuned on our dataset with \onestep/\twostep retrieval. For consistency, we use the same hyper-parameter sweep in all fine-tuning experiments (cf.\ Appendix~\ref{appendix:hyper}).

\subsection{Results on Crowd-Sourced Questions}

To enable fine-tuning models, we split the questions them into 5962/825/873 questions in train/dev/test folds, resp. To limit memorization, any two questions using the same seed fact, $\fS$, were always put in the same fold. Since multiple facts can cover similar topics, we further ensure that similar facts are also in the same fold. (See Appendix~\ref{appendix:datasplit} for details.)

While these crowd-sourced questions were challenging for the baseline QA models (by design), models fine-tuned on this dataset perform much better. The BERT baseline that scored 38.7\% on the crowd-sourced questions now scores 63.3\% on the dev set after fine-tuning. Even the basic \onestep retrieval context can improve over this baseline score by 14.9\% (score: 78.2\%) and our proposed \twostep retrieval improves it even further by 8.2\% (score: 86.4\%). This shows that the distractor choices selected by the crowdsource workers were not as challenging once the model is provided with the right context. This can be also seen in the incorrect answer choices selected by them in Table~\ref{tab:question-examples} where they used words such as ``Pain'' that are associated with words in the question but may not have a plausible  reasoning chain. To make this dataset more challenging for these models, we next introduce adversarial distractor choices.

\section{Adversarial Choice Generation}
\label{sec:adversarial}

To make the crowdsourced dataset challenging for fine-tuned language models, we use model-guided \emph{adversarial choice generation} to expand each crowdsourced question into an 8-way question. Importantly, the human-authored body of the question is left intact (only the choices are augmented), to avoid a system mechanically reverse-engineering how a question was generated.

\begin{figure}[t]
        \centering
        \includegraphics[width=0.4\textwidth]{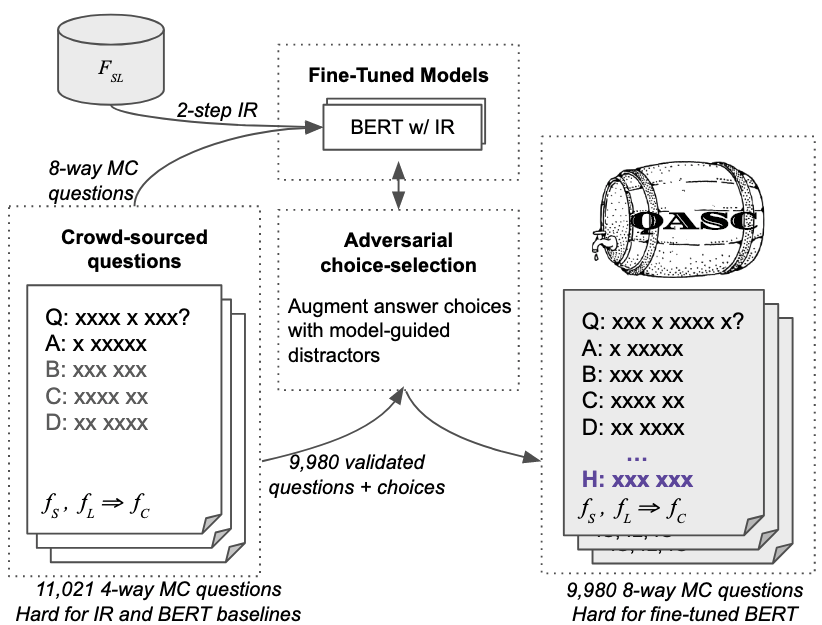}
        \caption{
            \label{fig:dataset-generation} Generating \dataset questions using adversarial choice selection.
        }
\end{figure}

Previous approaches to adversarially create a hard dataset have focused on iteratively making a dataset harder by sampling harder choices and training stronger models~\cite{dataset:swag,dataset:vcr}. While this strategy has been effective, it involves multiple iterations of model training that can be prohibitively expensive with large LMs. In some cases~\cite{dataset:swag,dataset:hellaswag}, they need a generative model such as GPT-2~\cite{gpt2} to produce the distractor choices. We, on the other hand, have a simpler setup where we train only a few models and do not require a model to generate the distractor choices.

\subsection{Distractor Options}

To create the space of distractors, we follow \namecite{dataset:vcr} and use correct answer choices from other questions. This ensures that a model won't be able to predict the correct answer purely based on the answer choices (one of the issues with OpenBookQA). To reduce the chances of a correct answer being added to the set of distractors, we pick them from the most dissimilar questions. We further filter these choices down to $\sim$30 distractor choices per question by removing the easy distractors based on the fine-tuned BERT baseline model.  Further implementation details are provided in Appendix~\ref{appendix:distractors}.

This approach of generating distractors has an additional benefit: we can recover the questions that were rejected earlier for having multiple valid answers (in \S~\ref{subsec:validation}). We add back 2,774 of the 3,361 rejected questions that (a) had at least one worker select the right answer, and (b) were deemed unanswerable by at most two workers. We, however, ignore all crowdsourced distractors for these questions since they were considered potentially correct answers in the validation task. We use the adversarial distractor selection process (to be described shortly) to add the remaining 7 answer choices. 

To ensure a clean evaluation set, we use \textbf{another crowdsourcing task} where we ask 3 annotators to identify \emph{all possible valid answers} from the candidate distractors for the dev and test sets. We filter out answer choices in the distractor set that were considered valid by at least one turker. Additionally, we filter out low-quality questions where more than four distractor choices were marked valid or the correct answer was not included in the selection. This dropped 20\% of the dev and test set questions and finally resulted in train/dev/test sets of size 8134/926/920 questions with an average of 30/26.9/26.1 answer choices (including the correct one) per question.

\begin{table}[t]
    \centering
    \setlength{\tabcolsep}{3pt}
    \small
    \begin{tabular}{lcc}
    \; & \multicolumn{2}{c}{Dev Accuracy} \\ \cline{2-3}
    \T \; & \Onestep retr. & \Twostep retr. \\
    \hline
    \T Original Dataset (4-way) & 78.2 & 86.4 \\
    Random Distractors (8-way) & 74.9 & 83.3 \\ 
    Adversarial Distractors (8-way) & 61.7 & 72.9 \\
    \end{tabular}
     \caption{
        \label{tab:crowdsourced_exp} Results of the \bertmcq model on the adversarial dataset using \texttt{bert-large-cased} model and pre-trained on \race + \sci questions.
        }
\end{table}

\begin{table*}[ht]
    \centering
    \small
    \setlength{\tabcolsep}{8pt}
    \begin{tabular}{clllll|cc}
    & & & Retr. Corpus & Retrieval & Addnl. fine-tuning & Dev & Test \\
    & Model & Embedding & (\#docs) & Approach & (\#examples) & Acc. & Acc. \\
    \hline
    \T & Human Score & &  &  & & & \textbf{93.0} \\
    \B & Random & & & & & 12.5 & 12.5 \\
    \hline
    \multirow{4}{*}{\rotatebox[origin=c]{90}{\parbox{1.1cm}{OBQA\\ Models}}}
     & ESIM Q2Choice & \glove & & & & 21.1 & 17.2 \\ 
    & ESIM Q2Choice & \glove \elmo & & & & 17.1 & 15.2 \\ 
    \B & Odd-one-out & \glove & & & & 22.4 & 18.0 \\ 
    \hline
    \multirow{4}{*}{\rotatebox[origin=c]{90}{\parbox{1.3cm}{BERT\\ Models}}}
    \T & BERT-MCQ & \bertlc & \FSL (17M) & \Onestep & & 59.8 & 53.2\\ 
    & BERT-MCQ & \bertlc & \FSL + ARC (31M) & \Onestep & & 62.3 & 57.0 \\ 
    & BERT-MCQ & \bertlc & \FSL + ARC(31M) & \Twostep & & 66.6 & 58.3 \\ 
    & BERT-MCQ & \bertlc & \FSL (17M) & \Twostep & & 71.0 & 67.0\\ 
    \hline
    \multirow{3}{*}{\rotatebox[origin=c]{90}{\parbox{1.0cm}{Addnl. \\ Fine-tuning}}}
     \T & AristoBertV7 & \bertwm & Aristo (1.7B) & \Onestep & \race + \sci (97K) & 69.5 & 62.6\\
    & BERT-MCQ & \bertlc & \FSL (17M) & \Twostep & \race + \sci (97K) & 72.9 & 68.5\\ 
    & BERT-MCQ & \bertwm & \FSL (17M) & \Twostep & \race + \sci (97K) & \textbf{78.0} & \textbf{73.2} \\ 
    \end{tabular}
    \caption{
        \label{tab:experiments}
       \dataset scores for previous state-of-the-art models on multi-hop Science MCQ(OBQA), and BERT models with different corpora, retrieval approaches and additional fine-tuning. While the simpler models only show a small increase relative to random guessing, BERT can achieve upto 67\% accuracy by fine-tuning on \dataset and using the \twostep retrieval. Using the BERT models pre-trained with whole-word masking and first fine-tuning on four relevant MCQ datasets (RACE and SCI(3)) improves the score to 73.2\%, leaving a gap of over 19.8\% to the human baseline of 93\%. ARC refers to the corpus of 14M sentences from \namecite{ARCClark2018}, \bertlc\ indicates `bert-large-cased` and \bertwm indicates whole-word masking.}
\end{table*}

\subsection{Multi-Adversary Choice Selection}

We first explain our approach, assuming access to $K$ models for multiple-choice QA. Given the number of datasets and models proposed for this task, this is not an unreasonable assumption. In this work, we use $K$ BERT models, but the approach is applicable to any QA model. 

Our approach aims to select a diverse set of answers that are challenging for different models. As described above, we first create $\sim$30 distractor options, $\disr$ for each question. We then sort these distractor options based on their relative difficulty for these models, defined as the number of models fooled by this distractor: $\sum_k \mathbf{I}\big[\mk(\ques, \disi) > \mk(\ques, \ans)\big]$ where $\mk(\ques, \choicei)$ is the $k$-th model's score for the question \ques and choice \choicei. In case of ties, we then sort these distractors based on the difference between the scores of the distractor choice and the correct answer: $\sum_k \big(\mk(\ques, \disi) - \mk(\ques, \ans)\big)$.\footnote{Since we use normalized probabilities as model scores, we do not normalize them here.}

We used \bertmcq models that were fine-tuned on the \race+\sci dataset as described in the previous section. We additionally fine-tune these models on the training questions with random answer choices added from the the space of distractors to make each question an 8-way multiple-choice question. This ensures that our models have seen answer choices from both the human-authored and algorithmically selected space of distractors. Drawing inspiration from bootstrapping~\cite{breiman1996bagging}, we create two such datasets with randomly selected distractors from $\disr$ and use the models fine-tuned on these datasets as $\mk$ (i.e, $K=2$). There is a large space of possible models and scoring functions that may be explored,\footnote{For example, we evaluated the impact of increasing $K$, but didn't notice any change in the fine-tuned model's score.} but we found this simple approach to be effective at identifying good distractors. This process of generating the adversarial dataset is depicted in Figure~\ref{fig:dataset-generation}.

\subsection{Evaluating Dataset Difficulty}
We select the top scoring distractors using the two \bertmcq models such that each question is converted into an 8-way MCQ (including the correct answer and human-authored valid distractors). To verify that this results in challenging questions, we again evaluate using the \bertmcq models with two different kinds of retrieval. Table~\ref{tab:crowdsourced_exp} compares the difficulty of the adversarial dataset to the original dataset and the dataset with random distractors (used for fine-tuning \bertmcq models).

The original 4-way MCQ dataset was almost solved by the \twostep retrieval approach and increasing it to 8-way with random distractors had almost no impact on the scores. But our adversarial choices drop the scores of the BERT model given context from either of the retrieval approaches.

\subsection{\dataset Dataset}
\label{sec:dataset}

\begin{table}[t]
    \centering
    \small
    \setlength{\tabcolsep}{5pt}
    \setlength{\doublerulesep}{\arrayrulewidth}
    \begin{tabular}{l|rrr}
    & Train & Dev & Test \\
    \hline \hline
    \T Number of questions & 8,134 & 926 & 920  \\
    Number of unique $\fS$ &   722 & 103 & 103 \\
    Number of unique $\fL$ & 6,157 & 753 & 762 \\
    Average question length (chars) & 46.4 & 45.5 & 44.0 \\
    \hline
    \end{tabular}
    \caption{
        \label{tab:dataset_desc}
        \dataset dataset statistics
    }
\end{table}

The final dataset contains \textbf{\datasetSize} questions split into \textbf{[8134$\mid$926$\mid$920]} questions in the [train$\mid$dev$\mid$test] folds. Each question is annotated with two facts that can be used to answer the question. These facts are present in a corpus of 17M sentences (also provided). 
The questions are similar to the examples in Table~\ref{tab:question-examples} but expanded to an 8-way MCQ and with shuffled answer choices. E.g., the second example there was changed to ``What forms caverns by seeping through rock and dissolving limestone? (A) pure oxygen (B) Something with a head, thorax, and abdomen (C) basic building blocks of life (D) \textbf{carbon dioxide in groundwater} (E) magma in groundwater (F) oxygen in groundwater (G) At the peak of a mountain (H) underground systems''. 

Table~\ref{tab:dataset_desc} gives a summary of \dataset statistics, and Table~\ref{tab:dataset-examples} in the Appendix provides additional examples.

\section{Experiments}

\label{sec:experiments}

While we used large pre-trained language models first fine-tuned on other QA datasets ($\sim$100K examples) to ensure that \dataset is challenging, we also evaluate BERT models without any additional fine-tuning here. All models are still fine-tuned on the \dataset dataset.

To verify that our dataset is challenging also for models that do not use BERT (or any other transformer-based architecture), we evaluate Glove~\cite{pennington2014glove} based models developed for multiple-choice science questions in OpenBookQA. Specifically, we consider these non-BERT baseline models:
 \begin{itemize}
     \item \textbf{Odd-one-out}: Answers the question based on just the choices by identifying the most dissimilar answer.
     \item \textbf{ESIM Q2Choice} (with and without Elmo): Uses the ESIM model~\cite{Chen2017EnhancedLF} with Elmo embeddings~\cite{elmo} to compute how much does the question entail each answer choice.
 \end{itemize}

As shown in Table~\ref{tab:experiments}, OpenBookQA models, that had close to the state-of-the-art results on OpenBookQA, perform close to the random baseline on \dataset. Since these mostly rely on statistical correlations between questions and across choices,\footnote{Their knowledge-based models do not scale to our corpus of 17M sentences.} this shows that this dataset doesn't have any easy shortcuts that can be exploited by these models.

Second, we evaluate BERT models with different corpora and retrieval. We show that our \twostep approach always out-performs the \onestep retrieval, even when given a larger corpus. Interestingly, when we compare the two \onestep retrieval models,  the larger corpus outperforms the smaller corpus, presumably because it increases the chances of having a single fact that answers the question. On the other hand, the smaller corpus is better for the \twostep retrieval approach,  as larger and noisier corpora are more likely to lead a 2-step search astray.

Finally, to compute the current gap to human performance, we consider a recent state-of-the-art model on multiple leaderboards: AristoBertV7 that uses the BERT model trained with whole-word masking,\footnote{https://github.com/google-research/bert} fine-tuned on the \race+\sci questions and retrieves knowledge from a very large corpus. Our \twostep retrieval based model outperforms this model and improves even further with more fine-tuning. Replacing the pre-trained \texttt{bert-large-cased} model with the whole-word masking based model further improves the score by 4.7\%,  but there is still a gap of $\sim$20\% to the human score of 93\% on this dataset.

\section{Conclusion}
\label{sec:conclusion}

We present \dataset, the first QA dataset for multi-hop reasoning beyond a single paragraph where two facts needed to answer a question are annotated for training, but questions cannot be easily syntactically decomposed into these facts. Instead, models must learn to retrieve and compose candidate pieces of knowledge. \dataset is generated via a crowdsourcing process, and further enhanced via multi-adversary distractor choice selection. State-of-the-art BERT models, even with massive fine-tuning on over 100K questions from previous relevant datasets and using our proposed \twostep retrieval, leave a large margin to human performance levels, thus making \dataset a new challenge for the community.

\subsection*{Acknowledgments}
\label{subsec:ack}

\begin{small}
We thank Oyvind Tafjord for his extension to AllenNLP that was used to train our BERT models, Nicholas Lourie for his ``A Mechanical Turk Interface (amti)'' tool used to launch crowdsourcing tasks, Dirk Groeneveld for his help collecting seed facts, and Sumithra Bhakthavatsalam for helping generate the QASC fact corpus. We thank Sumithra Bhakthavatsalam, Kaj Bostrom, Kyle Richardson, and Madeleine van Zuylen for initial human evaluations. We also thank Jonathan Borchardt and Dustin Schwenk for inspiring discussions about, and guidance with, early versions of the MTurk task. We  thank the Amazon Mechanical Turk workers for their effort in creating and annotating QASC questions. Computations on beaker.org were supported in part by credits from Google Cloud.
\end{small}

 {
 \fontsize{9.0pt}{10.0pt}
\selectfont
\bibliography{qasc}
\bibliographystyle{aaai}
 }

\clearpage
\appendix


\begin{table*}[htbp]
    \centering
    \small
    \setlength{\tabcolsep}{5pt}
    \setlength{\doublerulesep}{\arrayrulewidth}
    \begin{tabular}{p{6cm}p{3cm}p{3cm}p{3cm}}
    Question & Fact 1 & Fact 2 & Composed Fact \\
    \hline \hline
     \R What can trigger immune response? (A) decrease strength (B) \textbf{Transplanted organs} (C) desire (D) matter vibrating (E) death (F) pain (G) chemical weathering (H) an automobile engine    &  \hred{Antigens} are found on cancer cells and the cells of transplanted organs. & Anything that can trigger an immune response is called an \hred{antigen}. & transplanted organs can trigger an immune response \\
     \hline
    \R what makes the ground shake? (A) organisms and their habitat (B) \textbf{movement of tectonic plates} (C) stationary potential energy (D) It's inherited from genes (E) relationships to last distance (F) clouds (G) soil (H) the sun & an \hred{earthquake} causes the ground to shake & \hred{Earthquakes} are caused by movement of the tectonic plates. & movement of the tectonic plates causes the ground to shake \\
    \hline
    \R Salt in the water is something that must be adapted to by organisms who live in: (A) interact (B) condensing (C) aggression (D) repaired (E) Digestion (F) Deposition (G) \textbf{estuaries} (H) rainwater & Organisms that live in \hred{marine biomes} must be adapted to the salt in the water.	& Estuaries display characteristics of both \hred{marine} and freshwater \hred{biomes}.	& Organisms that live in estuaries must be adapted to the salt in the water. \\
    \hline
    \end{tabular}
    \caption{
        \label{tab:dataset-examples}
        Example of questions, facts and composed facts in \dataset. In the first question, the facts can be composed through the intermediate entity \facttext{antigen} to conclude that transplanted organs can trigger an immune response. 
    }
\end{table*}


\section{Appendix: Crowdsourcing Details}
\label{sec:crowdsourcing-details}

\subsection{Quality checking}

Each step in the crowdsourcing process was guarded with a check implemented on
the server-side. To illustrate how this worked, consider a worker eventually arriving 
at the following 4-way MC question:\\

\begin{tabular}{|l|l|}
  \hline
  Term     & Value \\
  \hline
  \fS      & \facttext{pesticides cause pollution} \\
  \fL      & \facttext{Air pollution harms animals} \\
  \fc      & \facttext{pesticides can harm animals} \\
  Question & \facttext{What can harm animals?} \\
  Answer A & \facttext{pesticides} \\
  Choice B & \facttext{manure} \\
  Choice C & \facttext{grain} \\
  Choice D & \facttext{hay} \\
  \hline
\end{tabular}\\

\noindent \textbf{Searching \FL}. The worker was presented with \fS \facttext{pesticides cause pollution} and asked to 
to search through \FL for a candidate \fL, which was compared to \fS by a quality checker to make sure it was appropriate. For example, searching for \facttext{pollution harms animals} would surface a candidate \fL \facttext{Tigers are fierce and harmful animals} which isn't admissible because it lacks overlap in significant words with \fS. On the other hand, the candidate \fL \facttext{Air pollution harms animals} is admissible by this constraint.\\

\noindent \textbf{Combining facts}. When combining \fS with \fL, the worker had to create a novel composition that omits significant words shared by \fS and \fL. For example, \facttext{pollutants can harm animals} is inadmissible because it mentions \facttext{pollutants}, which appears in both \fS and \fL. On the other hand, \facttext{pesticides can harm animals} is admissible by this constraint. After composing \fc, the worker had to highlight words in common between \fS and \fc, between \fL and \fc, and between \fS and \fL. The purpose of this exercise was to emphasize the linkages between the facts; if these linkages were difficult or impossible to form, the worker was advised to step back and compose a new \fc or choose a new \fL outright.\\

\noindent \textbf{Creating a question and answer}. After the worker proposed a question and answer, a quality check was performed to ensure both \fS and \fL were required to answer it. For example, given the question \facttext{What can harm animals} and answer \facttext{pollution}, the quality check would judge it as inadmissible because of the \textsl{bridge} word \facttext{pollution} shared by \fS and \fL that was previously removed when forming \fc. The answer \facttext{pesticides} does meet this criteria and is admissible.\\

\noindent \textbf{Creating distractor choices}. To evaluate proposed distractor choices, two QA systems were presented with the question, answer and the choice. The distractor choice was considered a distraction if the system preferred the choice over the correct answer, as measured by the internal mechanisms of that system. One of the systems was based on information retrieval (IR), and the other was based on BERT; these were playfully named \textsl{Irene AI} and \textsl{Bertram AI} in the interface.

\subsection{Crowdsourcing task architecture}

Creating each item in this dataset seemed overwhelming for crowd workers, and it wasn't clear how
to compensate for the entire work. Initially we presented two tasks
for independent workers. First, we collected a question and answer that could
be answered by combining a science fact with one more general fact. Second, we 
collected distractor choices for this question that fooled our systems.
However, the quality of submissions for both tasks was low. For the first task,
we believe that the simple checklist accompanying the task was insufficient to
prevent unreasonable submissions. For the second task, we believe the workers
were not in the mindset of the first worker, so the distractor choices too far
removed from the context of the question and its facts.

Ultimately we decided to combine the two simpler tasks into one longer and more
complex task. Additional server-side quality checks gave instant feedback in
each section, and numerous examples helped workers intuit the desire. We believe this
allowed workers to develop fluency to provide many submissions of a consistent
quality.

The crowd work was administered through Amazon Mechanical Turk (\url{https://www.mturk.com}).
Participating workers neeeded to be in the United States, hold a Master's qualification, have submitted 
thousands of HITs and had most of them approved. The work spanned several batches, each having its results checked before starting the next. See table ~\ref{table:mturk} for collection 
statistics.

\begin{table}[ht]
    \centering
    \small
    \setlength{\tabcolsep}{12pt}
    \setlength{\doublerulesep}{\arrayrulewidth}
    \begin{tabular}{l}
        \begin{tabular}{ccc}
            \hline
                   & Number of & Number of \\
            Batch  & questions & distinct \\
                   & collected & workers \\
            \hline \hline
            A      &    \ \ \ 376 &   17\\ 
            B      &    \ \ \ 759 &   23\\ 
            C      &    \ \ \ 872 &   21\\ 
            D      &   1,486 &   27\\ 
            E      &   1,902 &   26\\ 
            F      &   2,124 &   35\\ 
            G      &   1,877 &   38\\ 
            H      &   1,625 &  20 \\ 
            \hline  
            Total  &   11,021 &   62\\
            \hline
        \end{tabular} \\
    \end{tabular}
    \caption{
        \label{table:mturk}
        Questions collected. Batches A-F required workers to have completed 10,000 HITs and
        had 99\% of them accepted, while Batch G and H relaxed qualifications and
        required only 5,000 HITs with 95\% accepted.
    }
\end{table}

\subsection{User interface}

\begin{figure*}[ht] \centering
\fbox{ \includegraphics[width=0.85\textwidth,clip,trim=0cm 8.5cm 0cm 0.5cm]{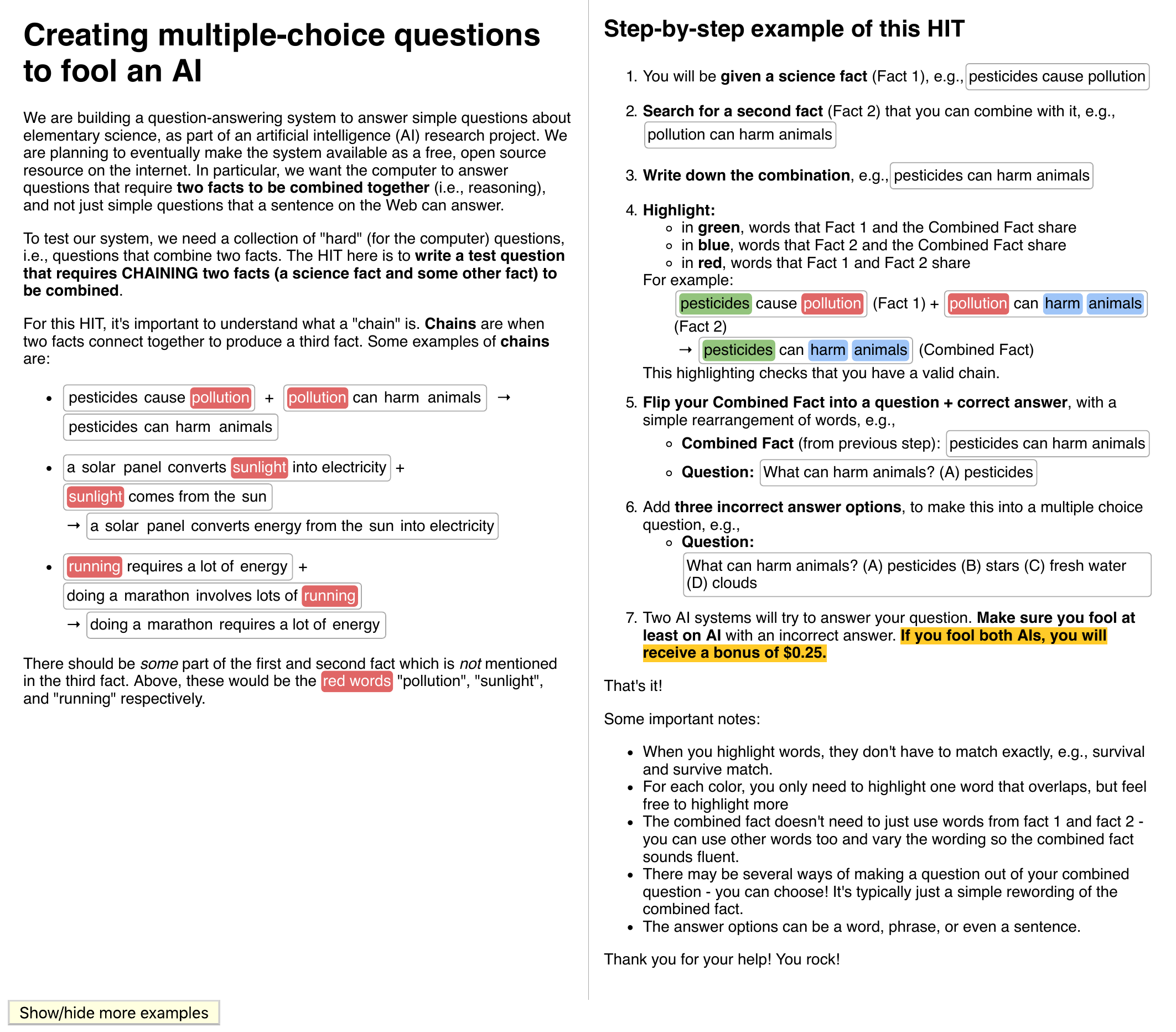} }
\caption{\label{fig:onebrain-1} MTurk HIT Instructions. Additional examples were revealed with a button.}
\end{figure*}

\begin{figure*}[ht] \centering
\fbox{ \includegraphics[width=0.85\textwidth,clip,trim=0cm 8.0cm 0cm 0.5cm]{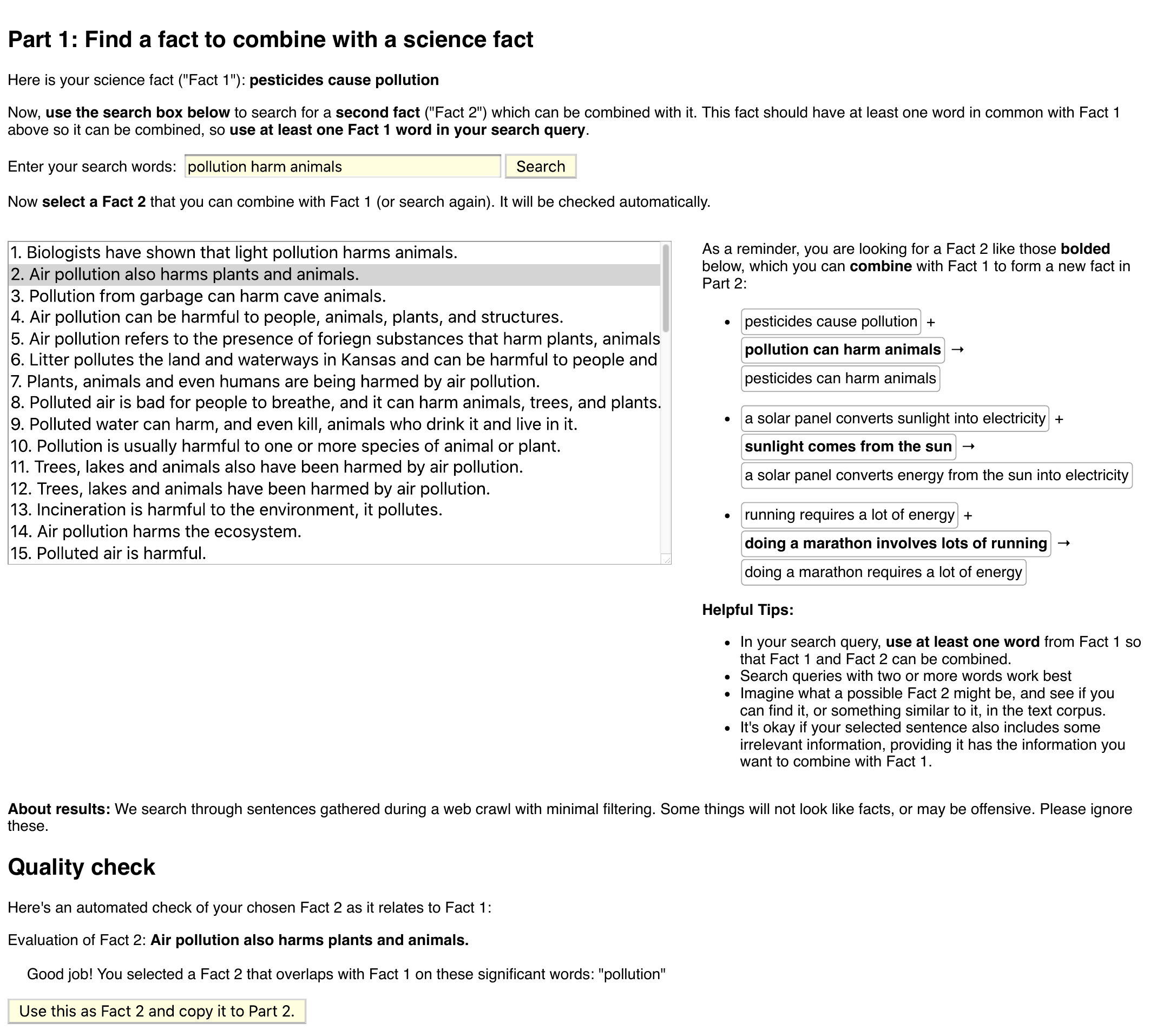} }
\caption{\label{fig:onebrain-2} MTurk HIT, finding \fL.}
\end{figure*}

\begin{figure*}[ht] \centering
\fbox{ \includegraphics[width=0.85\textwidth,clip,trim=0cm 11.5cm 0cm 1.0cm]{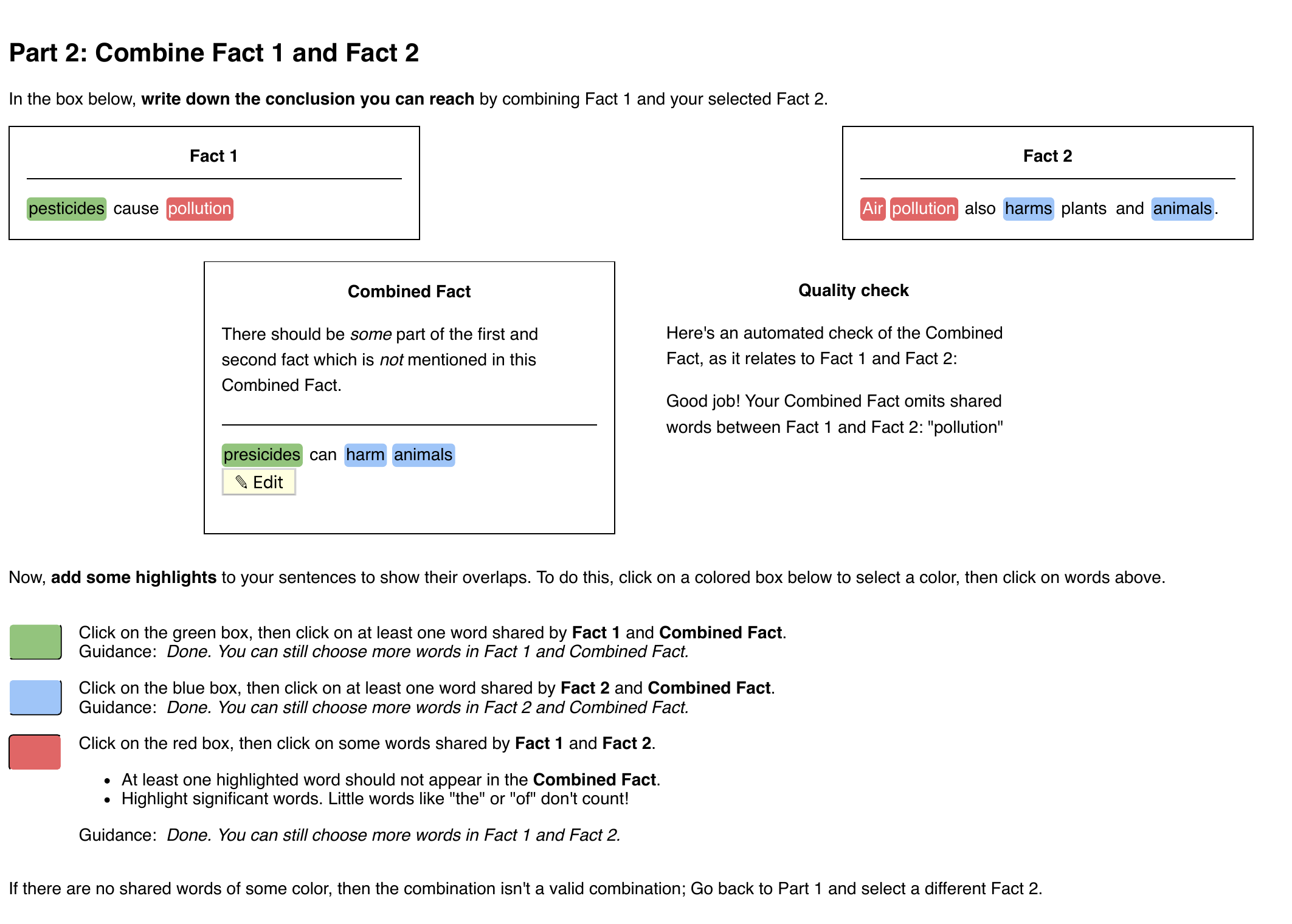} }
\caption{\label{fig:onebrain-3} MTurk HIT. Combining \fS and \fL to make \fc.}
\end{figure*}

\begin{figure*}[ht] \centering
\fbox{ \includegraphics[width=0.85\textwidth,clip,trim=0cm 2.0cm 0cm 1.0cm]{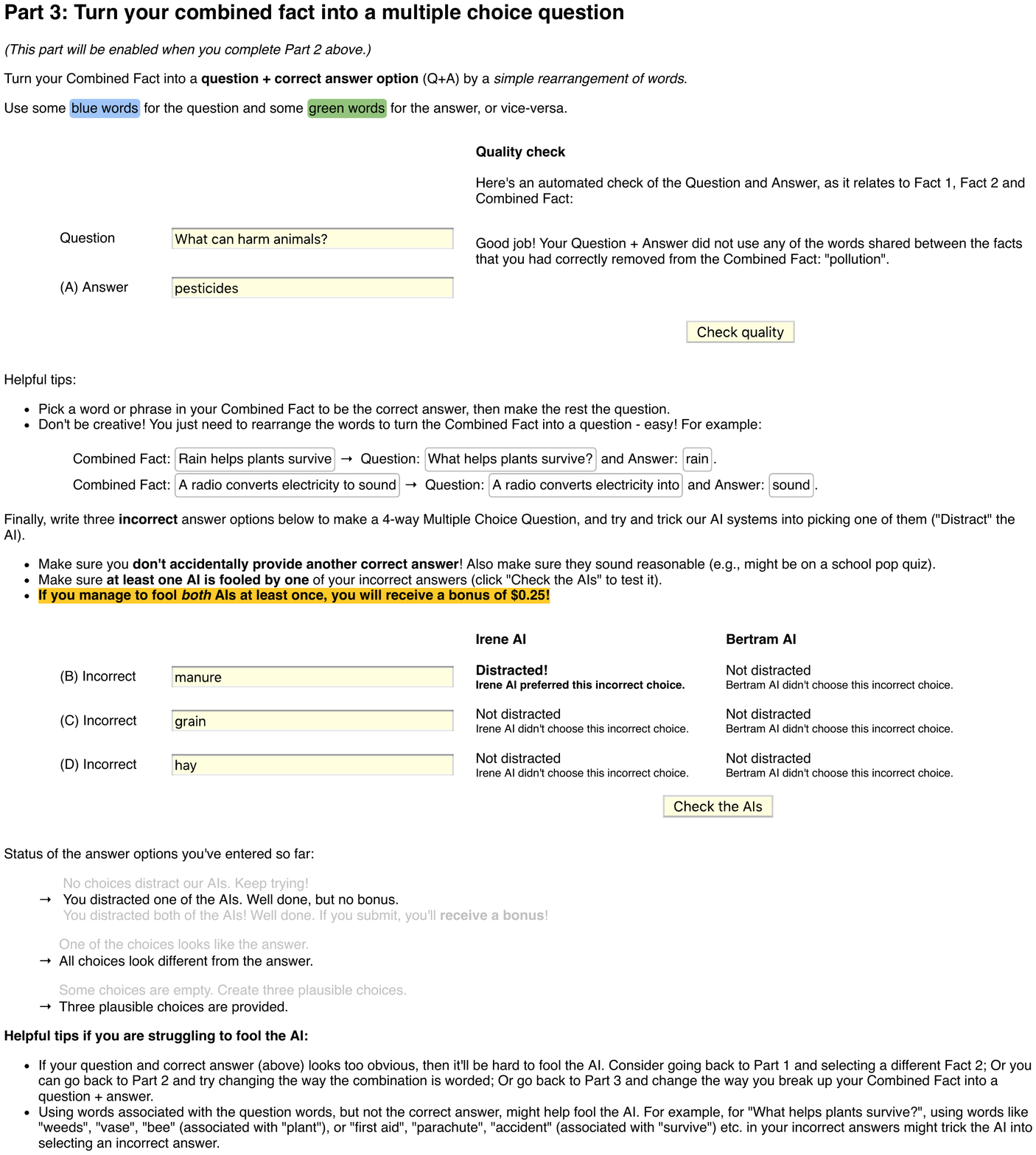} }
\caption{\label{fig:onebrain-4} MTurk HIT. Creating question, answer and distractor choices.}
\end{figure*}

Figures \ref{fig:onebrain-1}, \ref{fig:onebrain-2}, \ref{fig:onebrain-3}, and \ref{fig:onebrain-4} at the end of this document show the interface presented to crowd workers. Each step included guidance from external quality checkers and QA systems, so the workers could make progress only if specific criteria were satisfied.

Starting with the 928 seed facts in \FS, this process lead to 11,021 distinct human-generated questions.  Workers were compensated US\$1.25 for each question, and rewarded a bonus of US\$0.25 for each question that distracted both QA systems. On average workers were paid US\$15/hour. We next describe the baseline systems used in out process.

\section{Dataset Split}
\label{appendix:datasplit}
We use the seed fact \fS to compute the similarity between two questions. We use the tf-idf weighted overlap score (not normalized by length) between the two seed facts to compute this similarity score. Creating the train/dev/test split then can be viewed as finding the optimal split in this graph of connected seed facts. The optimality is defined as finding a split such that the similarity between any pair of seed facts in different folds is low. Additionally the train/dev/test splits should contain about 80/10/10\% of the questions. Note that each seed fact can have a variable number of associated questions.

Given these constraints, we defined a small Integer Linear Program to find the train/dev/test split. We defined a variable $n_{ij}$ for each seed fact $f_i$ being a part of each fold $t_j$, where only one of the variable per seed fact can be set to one ($\sum_j n_{ij} = 1$).  We defined an edge variable $e_{ik}$ between every pair of seed fact that is set to one, if the two facts $f_i$ and $f_k$ are in different folds. The objective function, to be minimized, is computed as the sum of the similarities between the facts in different folds i.e $\sum_{i,k} e_{ik} sim(f_i, f_k)$. Additionally, the train/dev/test fold were constrained to have 78/11/11\% of the questions (additional dev/test questions to account for the validation filtering step) with a slack of 1\% (e.g. for $t_j=$train, $0.77\times Q \leq \sum_i n_{ij} q_{i} \leq 0.79\times Q$ where $Q$ is the total number of questions and $q_i$ is the number of questions using $f_i$ as the seed fact.). To make the program more efficient, we ignore edges with a low tf-idf similarity (set to 10 in our experiments to ensure completion within an hour using GLPK).

\section{Distractor Options}
\label{appendix:distractors}
We create the candidate distractors from the correct answers of questions from the same fold. To ensure that most of these distractors are incorrect answers for the question, we pick the correct answers from questions (within the same fold) most dissimilar to this question. Rather than relying on the question's surface form (which does not capture topical similarity), we use the underlying source facts used to create the question. Intuitively, questions based on the fact \facttext{Metals conduct electricity} are more likely to have similar correct answers (metal conductors) compared to questions that use the word \facttext{metal}. Further, we restrict the answer choices to those that are similar in length to the correct answer, to ensure that a model cannot rely on text length to identify the correct answer.

We reverse sort the questions based on the word-overlap similarity between the two source facts and then use the correct answer choices from the top 300 most-dissmimilar questions. We only consider answer candidates that have at most two additional/fewer tokens and at most 50\% additional/fewer characters compared to the correct answer. We next use the BERT Baseline model used for the dataset creation, fine-tuned on random 8-way MCQ and evaluate it on all of the distractor answer choices. We pick the top 30 most distracting choices (i.e. highest scoring choices) to create the final set of distractors.

\section{Neural Model Hyperparameters}
\label{appendix:hyper}

\subsection{\Twostep Retrieval}
Each retrieval query is run against an ElasticSearch\footnote{https://www.elastic.co} index built over $\FSL$ with retrieved sentences filtered to reduce noise as described by \namecite{ARCClark2018}. The retrieval time of this approach linearly scales with the breadth of our search defined by K (set to 20 in our experiments). We set L to a small value (4 in our experiments) to promote diversity in our results and not let a single high-scoring fact from $F_1$ overwhelming the results.

\subsection{BERT models}
For BERT-Large models, we were only able to fit sequences of length 184 tokens in memory, especially when running on 8-way MC questions. We fixed the learning rate to 1e-5 as it generally performed the best on our datasets. We only varied the number of training epochs: \{4, 5\} and effective batch sizes: \{16, 32\}. We chose this small hyper-parameter sweep to ensure that each model was fine-tuned using the same hyper-paramater sweep while not being prohibitively expensive. Each model was selected based on the best validation set accuracy. 
We used the HuggingFace implementation\footnote{https://github.com/huggingface/pytorch-pretrained-BERT} of the BERT models within the AllenNLP~\cite{Gardner2017AllenNLP} repository in all our experiments. 


\section{Overlap Statistics}
\label{appendix:overlap_stats}

\begin{table}[ht]
    \centering
    \small
    \setlength{\tabcolsep}{15pt}
    \setlength{\doublerulesep}{\arrayrulewidth}
    \begin{tabular}{lr}
    & \% Questions \\
    min. sim(fact, \ques+\ans) $<$ 2 & 48.6 \\
    min. sim(fact, \ques+\ans) $<$ 3 & 82.5 \\
    min. sim(fact, \ques+\ans) $<$ 4 & 96.3 \\
    \hline
    & Avg. tokens \\
    sim(\ques + \ans, \fS) & 3.17 \\
    sim(\ques + \ans, \fL) & 1.98 \\
    \end{tabular}
     \caption{Overlap Statistics computed over the crowd-source questions. In the first three rows, we compute the \% of questions with at least one fact have $<k$ token overlap with the \ques+\ans. In the last two rows, we compute the average number of tokens that overlap with \ques+\ans from each fact. We use stop-word filtered stemmed tokens for these calculations.}
    \label{tab:dataset_overlap}
\end{table}


\end{document}